\documentclass[sort&compress, numafflabel]{elsarticle}

\usepackage[]{natbib}

\usepackage[breaklinks,hidelinks]{hyperref}
\usepackage{times}
\usepackage{latexsym}

\usepackage{geometry}

\usepackage{subfiles}
\usepackage{multirow}
\usepackage{array}
\usepackage{hyphenat}
\usepackage{subcaption}
\usepackage{longtable}
\usepackage{graphicx}
\usepackage{adjustbox}
\usepackage{enumitem}
\usepackage{url}

\usepackage[color=yellow, textsize=small]{todonotes}

\usepackage{bm}
\usepackage{booktabs}
\usepackage{float}
\usepackage[english]{babel}
\usepackage{blindtext}
\usepackage{textcomp}
\usepackage{soul,color}
\usepackage{pifont}

\makeatletter
\def\ps@pprintTitle{%
 \let\@oddhead\@empty
 \let\@evenhead\@empty
 \def\@oddfoot{}%
 \let\@evenfoot\@oddfoot}
\makeatother

\usepackage{epstopdf}
\AppendGraphicsExtensions{.tif}

\usepackage{titlesec}
\titleformat{\section}
      {\normalfont\bfseries}
      {\thesection}
      {0ex}
      {\MakeUppercase}

\titleformat{\subsection}
      {\normalfont\bfseries}
      {\thesection}
      {0ex}
      {}

\titleformat{\subsubsection}
      {\normalfont}
      {\thesection}
      {0ex}
      {}

\usepackage{setspace}
\newif\ifsubfile
\subfiletrue

\newif\iftif
\tiffalse

\usepackage{microtype}

\newcommand\blfootnote[1]{%
  \begingroup
  \renewcommand\thefootnote{}\footnote{#1}%
  \addtocounter{footnote}{-1}%
  \endgroup
}

\title{Leveraging Natural Language Processing to Augment Structured Social Determinants of Health Data in the Electronic Health Record}

\author[ist]{Kevin Lybarger\textsuperscript{*}\textsuperscript{+}\blfootnote{\textsuperscript{*}Authors contributed equally to this paper.}\blfootnote{\textsuperscript{+}Corresponding author: Kevin Lybarger, PhD, Department of Information Sciences and Technology, George Mason University, 4400 University Dr. MSN 1G8, Fairfax, VA 22030, USA; klybarge@gmu.edu,703-993-1547.}}

\ead{}
\author[bime,reIt]{Nicholas J Dobbins\textsuperscript{*}}
\ead{ndobb@uw.edu}
\author[reIt]{Ritche Long}
\ead{rlong07@uw.edu}
\author[med]{Angad Singh}
\ead{}
\author[med]{Patrick Wedgeworth}
\ead{}
\author[ist]{Özlem Uzuner}
\ead{}
\author[bime]{Meliha Yetisgen}
\ead{melihay@uw.edu}

\address[ist]{Department of Information Sciences and Technology, George Mason University, Fairfax, VA, USA \blfootnote{ \\ Word count: 4500} }
\address[bime]{Department of Biomedical Informatics \& Medical Education, University of Washington, Seattle, WA, USA}
\address[reIt]{Department of Research IT, UW Medicine, University of Washington, Seattle, WA, USA}
\address[med]{Department of Medicine, University of Washington, Seattle, WA, USA}

\begin{document}
\subfilefalse

\newpageafter{author}

\begin{abstract}

\noindent\textbf{Objective:} Social determinants of health (SDOH) impact health outcomes and are documented in the electronic health record (EHR) through structured data and unstructured clinical notes. However, clinical notes often contain more comprehensive SDOH information, detailing aspects such as status, severity, and temporality. This work has two primary objectives: i) develop a natural language processing (NLP) information extraction model to capture detailed SDOH information and ii) evaluate the information gain achieved by applying the SDOH extractor to clinical narratives and combining the extracted representations with existing structured data.  \\

\noindent\textbf{Materials and Methods:} We developed a novel SDOH extractor using a deep learning entity and relation extraction architecture to characterize SDOH across various dimensions. In an EHR case study, we applied the SDOH extractor to a large clinical data set with 225,089 patients and 430,406 notes with social history sections and compared the extracted SDOH information with existing structured data. \\

\noindent\textbf{Results:} The SDOH extractor achieved 0.86 F1 on a withheld test set. In the EHR case study, we found extracted SDOH information complements existing structured data with 32\% of homeless patients, 19\% of current tobacco users, and 10\% of drug users only having these health risk factors documented in the clinical narrative.   \\

\noindent\textbf{Conclusions:} Utilizing EHR data to identify SDOH health risk factors and social needs may improve patient care and outcomes. Semantic representations of text-encoded SDOH information can augment existing structured data, and this more comprehensive SDOH representation can assist health systems in identifying and addressing these social needs.

\end{abstract}


\maketitle








\pagebreak

\section*{Introduction}

Social determinants of health (SDOH) are increasingly recognized for their influence on patient health, accounting for an estimated 40\%-90\% of health outcomes.\cite{friedman2018toward} SDOH include \textit{protective factors} that reduce health risks (e.g. family support) and \textit{risk factors} that increase health risks (e.g. housing instability).\cite{alderwick2019meanings} SDOH interventions, such as initiating medication assisted therapy in opioid patients, demonstrate a clear reduction in mortality.\cite{ma2019effects} Other studies have demonstrated the importance of SDOH data in improving the prediction of hospital readmissions, medication adherence, suicide attempts, and more.\cite{nijhawan2019clinical, chen2020social} Such studies reinforce the importance of screening patients for social needs, so clinical care teams can connect them with needed resources.

Patient SDOH information is captured in the Electronic Health Record (EHR) through structured data and unstructured clinical narrative text. The clinical narrative contains a more nuanced and detailed representation of many SDOH than is available through structured data. For example, substance use (alcohol, tobacco, and drug) is often documented through binary fields (yes/no) in structured data, while clinical narratives often document substance use frequency, amount, and history information. This information can be automatically extracted using natural language processing (NLP) information extraction techniques, which map the unstructured text to a structured SDOH representation. Combining extracted  information from clinical narratives with existing structured data yields a more complete patient representation.\citep{navathe2018hospital, hatef2019assessing} This more complete, automatically-derived patient representation can be used in large-scale secondary use applications, including clinical decision-support systems and retrospective studies. Utilizing already-collected data may reduce the workload and financial resources required for data collection.

\ifsubfile
\bibliography{mybib}
\fi

\section*{Background and significance}

Secondary use of SDOH information from clinical narratives requires extraction of relevant information and conversion of the SDOH descriptions to structured semantic representations. The extraction of SDOH information from clinical text is increasingly explored; however, the nature and granularity of the target SDOH varies across the research space.\citep{patra2021extracting} Several studies treated SDOH extraction as a text classification task, where labels are assigned at the sentence or note-level.\citep{uzuner2008identifying, stemerman2021identification, gehrmann2018comparing,feller2018towards, yu2021study, han2022classifying} Narrative SDOH information has also been the target of relation or event extraction, where SDOH are characterized across multiple dimensions related to status, temporality, and other attributes.\citep{wang2016investigating, Yetisgen2017substance, LYBARGER2021103631} SDOH information extraction techniques include rule-based\citep{hatef2019assessing,lowery2022using,reeves2021adaptation} and supervised learning approaches such as Support Vector Machines, random forest, and logistic regression.\citep{patra2021extracting} More recent supervised extraction approaches utilized deep learning architectures, such as convolutional neural networks, recurrent neural networks, and pre-trained transformers, including Bidirectional Encoder Representations from Transformers (BERT)\cite{devlin2019bert} and Text-To-Text Transfer Transformer (T5)\cite{raffel2020exploring}.\citep{patra2021extracting, han2022classifying, lybarger2023n2c2} Pre-trained transformers, such as BERT and T5, allow pre-training on large quantities of unlabeled text and fine-tuning model parameters to specific classification tasks.\cite{devlin2019bert, raffel2020exploring} The fine-tuning of pre-trained transformers is an effective transfer learning strategy that has achieved state-of-the-art performance in several SDOH information extraction tasks.\citep{yu2021study, han2022classifying, lybarger2023n2c2}

Prior studies have applied SDOH extractors to clinical data sets to understand the prevalence of SDOH information within clinical narratives and assess information gained relative to existing structured data. Hatef et al. developed hand-crafted linguistic patterns for social isolation, housing insecurity, and financial strain, which were applied to a large clinical data set.\citep{hatef2019assessing} Navathe et al. used a rule-based system\citep{zhou2011using} to extract SDOH from notes and demonstrated a more complete representation of patient substance use, depression, housing instability, fall risk, and poor social support can be obtained when combined with diagnosis codes.\citep{navathe2018hospital} Zhang et. al similarly combined narrative text and structured data to predict patient outcomes using deep learning.\cite{zhang2020combining} Focusing on lung cancer patients, Yu, et al. utilized BERT and RoBERTa \cite{liu2019roberta} to identify SDOH concepts at the document-level and compared the extracted results with structured EHR data.\cite{yu2021study, yu2022assessing}

\subsection*{Contributions}

This article presents two main contributions. First, we present a state-of-the-art event-based deep learning extractor for SDOH, the multi-label span-based entity and relation transformer (mSpERT). mSpERT was trained on the Social History Annotated Corpus (SHAC),\cite{LYBARGER2021103631} the benchmark gold standard dataset from the 2022 National NLP Clinical Challenges SDOH extraction task (n2c2/UW SDOH Challenge).\cite{lybarger2023n2c2} In prior work, we developed SHAC to address the limitations of published studies in terms of SDOH representation and normalization. For example, Han et al. used BERT for sentence-level SDOH classification but did not extract granular information related to substance types, duration, and frequency.\cite{han2022classifying} Yu et al. extracted text spans referring to smoking but did not identify specific entity types or normalize the spans to SDOH concepts (e.g., the phrase ``smoked 2 packs per day until 5 years ago'' would not be labeled as a past habit or as containing specific frequency or amount information). SHAC is novel in the granularity of the annotations, size of the corpus, and inclusion of multi-institution data. The granular SDOH annotations enable a broader set of secondary downstream applications. If compared to n2c2/UW SDOH Challenge shared task systems trained and evaluated on SHAC, the mSpERT performance of 0.86 overall F1 would only be surpassed by two teams, which achieved 0.89 and 0.88 overall F1.\cite{lybarger2023n2c2} We provide the code, trained extraction model, and annotated data (SHAC)\footnote{https://github.com/uw-bionlp/mspert}. To our knowledge, this is the first publicly available SDOH extractor trained on SHAC to the research community. 

Our second contribution is a large scale EHR case study that demonstrates the utility of NLP for SDOH extraction. We measured the prevalence of substance use, living situation, and employment information in the clinical narrative and structured SDOH data. Previous studies exploring the prevalence of SDOH information in the clinical narrative are limited by the EHR dataset size, patient population scope, and extraction methods. These studies have either applied extractors to relatively small, often disease-specific  cohorts\cite{wang2016investigating,yu2021study,navathe2018hospital,stemerman2021identification, conway2019moonstone} or used rule-based  approaches.\cite{hatef2019assessing,navathe2018hospital, conway2019moonstone} In contrast, we applied  mSpERT to a large clinical dataset of 225,089 patents and 430,406 notes spanning all patient populations from University of Washington Medicine. We compared extracted SDOH information from clinical narratives with the structured EHR data. The results show that combining the narrative SDOH information with the existing structured data yields a more comprehensive patient representation, which can help guide patient care, assess health risks, and identify social needs.

\ifsubfile
\bibliography{mybib}
\fi

\section*{Materials and Methods}

To extract detailed representations of SDOH from the clinical narrative, we developed a high performing event-based SDOH extractor, mSpERT, using SHAC. mSpERT can extract multiple SDOH events in the patient timeline, including past and current SDOH, and characterize SDOH events through detailed arguments related to status, severity, type, and temporality. Through an EHR case study, we applied mSpERT to a University of Washington (UW) dataset that includes 430,406 notes with social history sections for 225,089 patients and compared the extracted information with existing structured data to quantify differences in SDOH coverage. The structured data captures SDOH information through coarse encounter-level labels. To facilitate a direct comparison between the extracted information and existing structured data, we mapped a subset of the extracted SDOH information to note-level labels. To validate mSpERT on the UW dataset, we randomly sampled and annotated a subset of the notes in the UW dataset with note-level labels that can be directly compared with structured fields. All parts of this work were approved by our institution's IRB. This section presents the: i) data used, ii) information extraction methodology, and iii) EHR case study design.

\subsection*{Data}

 Figure 1
 \begin{figure*}[ht]
   \begin{center}
     \includegraphics[width=6.0in]{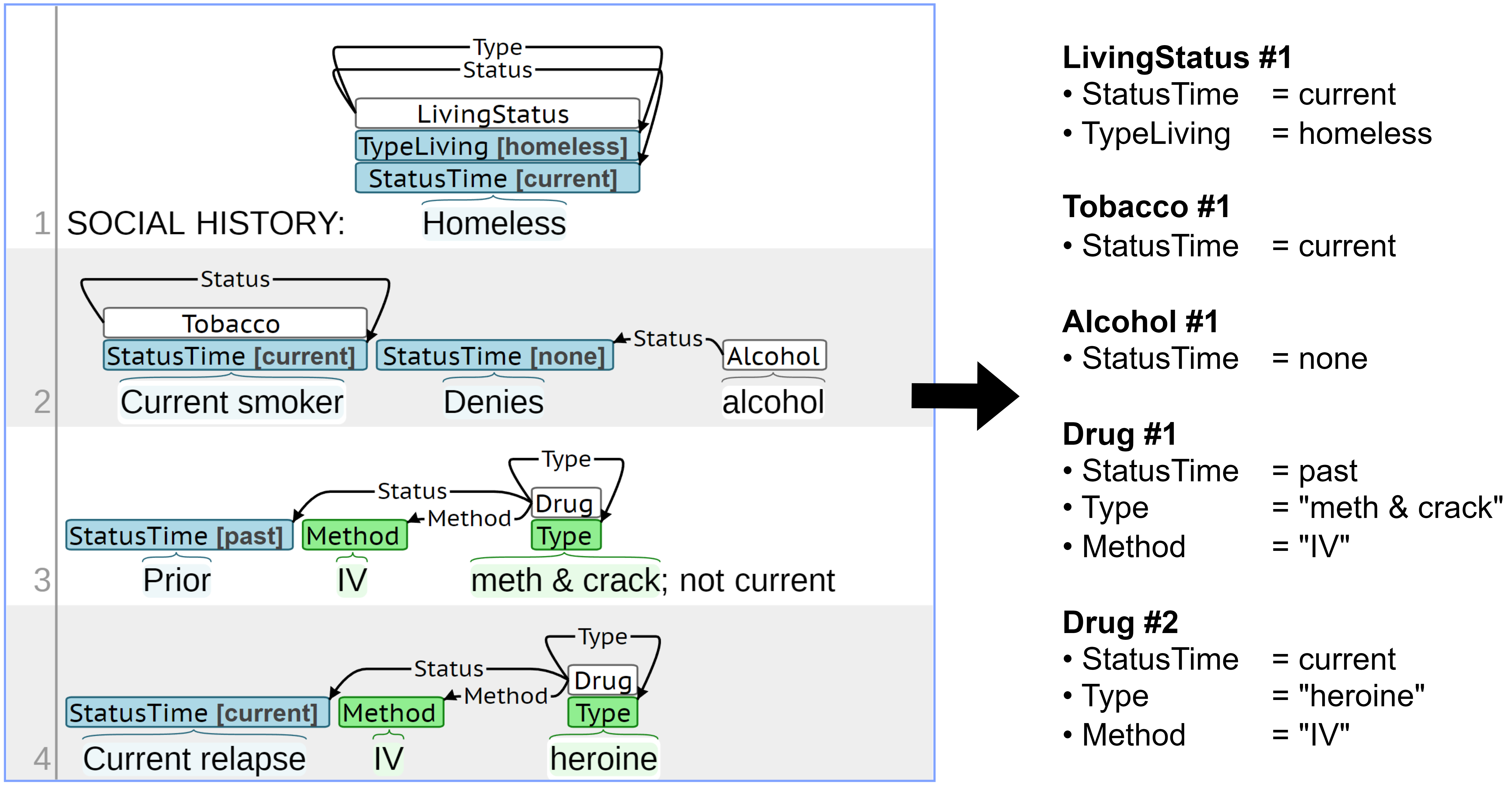}
   \end{center}
   \caption{SHAC annotation example (left side) with slot filling interpretation of the annotated events (right side).}
   \label{annotation_example}
 \end{figure*}

We used SHAC to develop and evaluate mSpERT.\cite{LYBARGER2021103631, lybarger2023n2c2} SHAC includes 4,405 annotated social history sections from clinical notes from MIMIC-III\citep{johnson2016mimic} and UW. It includes train, development, and test partitions for both sources. SHAC uses an event-based schema, where each event includes a trigger that identifies the event type and arguments that characterize the event. The SHAC event annotation schema characterizes each SDOH event in the patient timeline across multiple dimensions. Figure 1 presents an annotated social history section from SHAC with a slot filling interpretation of the events. In this example, the event schema can differentiate between the current use of heroin and the past use of methamphetamines and crack. It can also resolve the method of use, intravenous, for all substances. The slot filling representation in Figure 1 illustrates how the SDOH annotation scheme can be mapped to a structured format for utilization in secondary use applications. Table \ref{annotated_phenomena} summarizes the arguments for each event type: \textit{Alcohol}, \textit{Drug}, \textit{Tobacco}, \textit{Employment}, and \textit{Living Status}. There are two categories of arguments: i) \textit{span-only} arguments (green labels in Figure 1), which include an annotated span (e.g. ``IV'') and argument type (e.g. \textit{Method}) and ii) \textit{labeled} arguments (blue labels in Figure 1), which include an annotated span (e.g. ``Prior''), argument type (e.g. \textit{Status Time}), and argument subtype (e.g. \textit{past}) that normalize the span to key SDOH concepts. The argument subtype labels associated with the labeled arguments provide discrete features for downstream applications and improve the utility of the extracted information. Additional information regarding SHAC, including the distribution of note counts by source and annotation details, is available in the original SHAC paper and the n2c2/UW SDOH Challenge paper.\cite{LYBARGER2021103631, lybarger2023n2c2}


\begin{table}[ht]
    \small
    \centering
    \begin{tabular}{p{0.7in} p{0.8in} p{2.2in} p{1.2in}}
\toprule
\textbf{Event type}                                         & \textbf{Argument type} & \textbf{Argument subtypes}                                     & \textbf{Span examples}                                            \\ \midrule
\multirow{8}{0.8in}[\baselineskip]{Alcohol, Drug, \& Tobacco} & Status Time\textsuperscript{*}            & \{none, current, past\}                   & ``drinks," ``reports"                             \\ \cline{2-4} 
                                                            & Duration          & --                                                                  & ``for 10 years"                                            \\ \cline{2-4} 
                                                            & History           & --                                                                  & ``2 years ago"                                                 \\ \cline{2-4} 
                                                            & Type              & --                                                                  & ``whiskey," ``meth"                                         \\ \cline{2-4} 
                                                            & Amount            & --                                                                  & ``1-2 drinks," ``1 pack"                                           \\ \cline{2-4} 
                                                            & Frequency         & --                                                                  & ``a day," ``weekly"                                      \\ \hline 
\multirow{4}{0.8in}[\baselineskip]{Employment}              & Status Employ\textsuperscript{*}            & \{\nohyphens{employed, unemployed, retired, \newline on disability, student, homemaker}\} & ``working," ``retired"               \\ \cline{2-4} 
                                                            & Duration          & --                                                                  & ``for 15 years"                                         \\ \cline{2-4} 
                                                            & History           & --                                                                  & ``last year"                                                    \\ \cline{2-4} 
                                                            & Type              & --                                                                  & ``construction," ``lawyer"                                 \\ \hline
\multirow{4}{0.8in}[\baselineskip]{Living Status}                          & Status Time\textsuperscript{*}            & \{current, past, future\}    & ``living," ``resides"                                         \\ \cline{2-4} 
                                                            & Type Living\textsuperscript{*}              & \{\nohyphens{alone, with family, with others, homeless}\}  & ``with family," ``homeless"                        \\ \cline{2-4} 
                                                            & Duration          & --                                                                  & ``for 2 years"                                         \\ \cline{2-4} 
                                                            & History           & --                                                                  & ``until last year"                                               \\ \bottomrule

\end{tabular}
    \caption{Annotation guideline summary. *indicates a labeled argument. The labeled arguments are required for each event.}
    \label{annotated_phenomena}
\end{table}

To assess the SDOH coverage in the clinical narrative relative to structured data, we created a clinical dataset from  UW from January 1-December 31, 2021, which we refer to as the \textit{UW Data Set}. The UW Data Set includes structured and narrative text data from the UW Epic EHR from outpatient, emergency, and inpatient settings, including 20 medical specialities. Table \ref{tbl_datasets} summarizes the total records and unique patients. UW Data Set contained more than 3.3 million notes for 225,089 patients. In the case study, we processed 430,406 notes with social history sections with mSpeRT. To validate mSpERT, we created the \textit{UW Validation Set}, which consists of 750 randomly sampled documents with social history sections with equal proportions of progress notes, emergency notes, and social history documents. 

\begin{table}[!ht]
    \small
    \centering
    \def\arraystretch{1.3}
\begin{tabular}{m{1.7cm} l c c c c}
 \toprule
 
 \textbf{Data Type} & \textbf{Name} & \textbf{Total Records} & \textbf{Total Records with Social History} & \textbf{Unique Patients} \\
    \hline
    \multirow{5}{*}{\mbox{Structured}}
    & Flowsheets & 83,235 & -- & 7,875 \\
    & Social History & 733,591 & -- & 297,581 \\
    & Occupation History & 120,733 & -- & 42,115 \\
    & Employment Status & 560,940 & -- & 560,940 \\
    & \textbf{Total} & \textbf{1,498,499} & -- & \textbf{618,363} \\
    \hline
    \multirow{4}{*}{\mbox{Free-Text}}
    & Progress Notes & 3,063,025 & 283,423 & 140,820 \\
    & ED Notes & 147,114 & 19,120 & 14,619  \\
    & Social History Doc. & 127,863 & 127,863 & 127,863 \\
    & \textbf{Total} & \textbf{3,338,002} & \textbf{430,406} & \textbf{225,089} \\ \bottomrule
    
\end{tabular}
    \caption{Data sources used in the EHR case study. ``Total Records'' indicates the total counts of structured data records and free-text documents. ``Total Records with Social History'' indicates the number of progress and emergency (ED) notes with social history sections and number of social history entries.}
    \label{tbl_datasets}
\end{table}

\subsection*{Information Extraction}

\subsubsection*{Event Extraction}

The SHAC events can be decomposed into a set of relations, where the head is the trigger, tail is an argument, and relation type is the argument role. To extract SHAC events, we introduce mSpERT, which builds on Eberts and Ulges's SpERT.\cite{eberts2020span} SpERT jointly extracts entities and relations using BERT \cite{devlin2019bert} with output layers that classify spans and predict span relations. SpERT achieved state-of-the-art performance in multiple extraction tasks.\cite{eberts2020span}
SpERT's span-based architecture allows overlapping span predictions but only allows a single label to be assigned to each span; however, the SHAC annotations frequently assign multiple labels to a single span. To adapt SpERT to SHAC, we developed mSpERT. Figure 2 presents the mSpERT framework, which includes three classification layers: 1) Entity Type, 2) Entity Subtype, and 3) Relation. The input is a sentence, and the output is extracted events. The Entity Type and Relation layers are identical to the original SpERT, and the Entity Subtype layer is incorporated to generate multi-label span predictions.  

 Figure 2
 \begin{figure*}[ht]
   \begin{center}
     \includegraphics[width=6.0in]{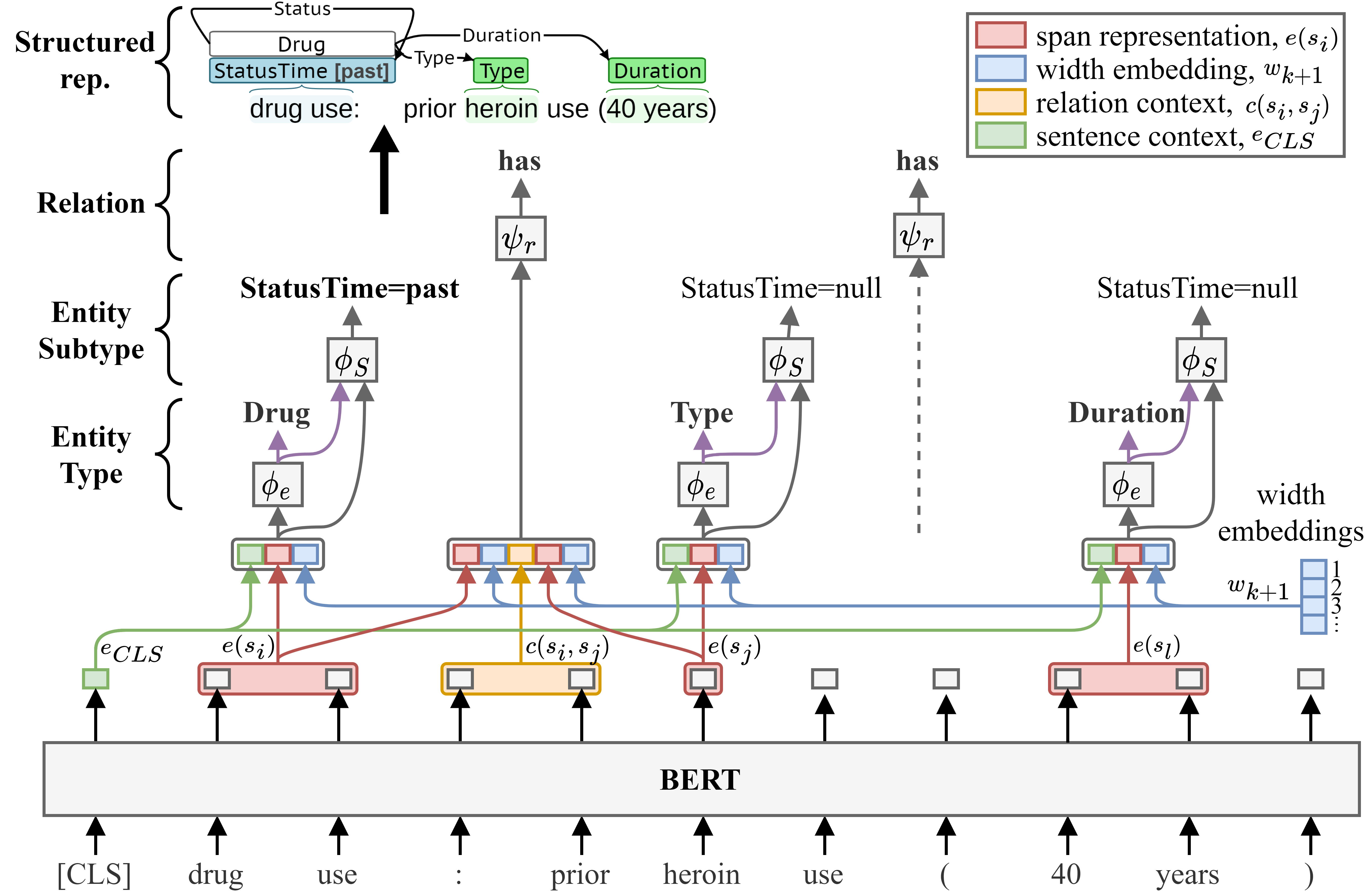}
   \end{center}
   \caption{Multi-label Span-based Entity and Relation Transformer (mSpERT) model, which builds on the original SpERT framework.\cite{eberts2020span}}
   \label{mspert}
 \end{figure*}

\textbf{Input encoding:} BERT generates a sequence of word-piece embeddings $(\bm{h}_{CLS}, \bm{h}_1, ... \bm{h}_t,..., \bm{h}_n)$, where $\bm{h}_{CLS}$ is the sentence representation, $\bm{h}_t$ is the $t^{th}$ word piece embedding, and $n$ is the sequence length.

\textbf{Entity Type:} The Entity Type classifier labels each span, $s_i = (t, {t+1}, ...{t+k})$, where $i$ is the span index and $k+1$ is the span width. Learned span width embeddings, $\bm{w}$, incorporate a span width prior. The span representation, $g(s_i)$, is generated from the BERT embeddings of $s_i$ and the width embeddings, as
\begin{equation}
g(s_i) = MaxPool(\bm{h}_t, \bm{h}_{t+1}, ...\bm{h}_{t+k})\circ\bm{w}_{k+1},
\end{equation}
where $\circ$ denotes concatenation. 
The Entity Type classifier is a linear layer, $\phi_e$, operating on $\bm{x}_{s,i}$, defined as
\begin{equation}
\bm{x}_{s,i} = g(s_i)\circ\bm{h}_{CLS}.
\label{span_classifier_input}
\end{equation}

\textbf{Entity Subtype:} The Entity Subtype classifiers consist of separate linear layers, $\phi_{s, v}$, where $v$ indicates the argument type. The Entity Subtype classifiers operate on the same span representation as the Entity Type classifier and incorporate the Entity Type classifier logits, as
\begin{equation}
\bm{x}_{s, v, i} = \bm{x}_{s,i}\circ\phi_e(\bm{x}_{s,i}).
\label{subtype_prediction_equation}
\end{equation}
The Entity Type logits are incorporated to improve the consistency between entity type and subtype predictions. 

\textbf{Relation:} The Relation classifier predicts the relationship between a candidate head span, $s_i$, and a candidate tail span, $s_j$, with input 
\begin{equation}
\bm{x}_{r, i,j} = g(s_i)\circ\bm{c}(s_i, s_j)\circ g(s_j),
\end{equation}
where $g(s_i)$ and $g(s_j)$ are the head and tail span embeddings and $\bm{c}(s_i, s_j)$ is the max pooling of the embedding sequence between the head and tail spans. The Relation classifier consists of a linear layer, $\phi_r$.

The Entity Type label set, $\Phi_e$, includes the $null$ label, event types (\textit{Alcohol}, \textit{Drug}, \textit{Tobacco}, \textit{Employment}, and \textit{Living Situation}), and span-only arguments (\textit{Amount}, \textit{Duration}, \textit{Frequency}, \textit{History}, and \textit{Type}) ($|\Phi_e|=11$). For all classifiers, $null$ is the negative label. There are three Entity Subtype classifiers (\textit{Status Time}, \textit{Status Employ}, and \textit{Type Living}), and the label set for each classifier includes $null$ and the applicable subtype labels (e.g. $\{null, none, current, past\}$ for \textit{Alcohol}). In SHAC, the links between the arguments and triggers can be interpreted as binary connectors (\textit{has} vs. \textit{does not have}). Consequently, the Relation label set, $\Phi_r$, is $\{null, has\}$. Only spans predicted to have a non-$null$ label by the Entity Type classifier are considered in relation classification. 

\textbf{Training:} The classification layers were learned while fine-tuning BERT. 
The training spans include all the gold spans, $S^g$, as positive examples and a fixed number of spans with label $null$ as negative examples. The training relations include all the gold relations as positive samples, and negative relation examples are created from entity pairs in $S^g$ not connected through a relation. Hyperparameters were tuned using the SHAC training and development sets, and final performance was assessed on the UW partition of the withheld SHAC test set. 

\textbf{Evaluation:} We used the n2c2/UW SDOH Challenge evaluation criteria, which interprets event extraction as a slot filling task.\cite{lybarger2023n2c2} In secondary use, there may be multiple semantically similar annotations, and the evaluation uses relaxed criteria that reflect the clinical meaning of the extractions.

\textit{Trigger:} A trigger is defined by an event type and multi-word span. Trigger equivalence is defined using \textit{any overlap} criteria where triggers are equivalent if: 1) the event types match and 2) the spans overlap by at least one character. 

\textit{Arguments:} Events are aligned based on trigger equivalence, and the arguments of aligned events are compared using different criteria for \textit{span-only} and \textit{labeled} arguments. 

\textit{Span-only arguments:}  A span-only argument is defined by an argument type, argument span, and trigger connection. Span-only argument equivalence is defined using \textit{exact match} criteria; span-only arguments are equivalent if: 1) the connected triggers are equivalent, 2) the argument types match, and 3) the spans match exactly.

\textit{Labeled arguments:} A labeled argument is defined by an argument type, argument subtype, argument span, and trigger connection. Label arguments are defined using a \textit{span agnostic} approach, where labeled arguments are equivalent if: 1) the connected triggers are equivalent, 2) the argument types match, and 3) the argument subtypes match. The argument span is not considered, and the span of the connected trigger is used as a proxy for argument location. 

Extraction performance was evaluated using the SHAC gold standard labels. A more detailed description of the scoring criteria and its justification is available in the n2c2/UW SDOH Challenge paper.\cite{lybarger2023n2c2}

\subsection*{EHR Case Study}


Our EHR case study consisted of two experiments: i) we validated mSpERT on the UW Data Set using 750 human-annotated documents, called the UW Validation Notes, and ii) we identified 1.4 million SDOH-related structured records for 618,363 patients and compared directly with NLP-derived data from 430,406 documents with social history sections for 225,089 patients written in 2021. The existing structured data did not capture the same granularity of SDOH information as mSpERT, so we mapped the mSpERT output to note-level labels that can be directly compared with structured fields.

\subsubsection*{Data Sources}

\textbf{Structured Data}: In this study, we used four database tables in the UW Epic EHR: flowsheets, social history, patient employment status, and patient occupation. From the flowsheets table, we leveraged the SDOH-related records identified by Phuong, et al. \cite{phuong2021extracting} to identify employment and housing status. The social history table is primarily composed of Boolean yes/no columns related to alcohol, tobacco, illicit, and recreational drug use. The patient employment status table provides current categorical employment status, such as $Student$, $Full-time$, and $Retired$, while the patient occupation table provides a longitudinal record of free-text occupation titles, such as ``Mechanic'' or ``Therapist.'' The employment status table does not include timestamps to determine when records were updated, so we limited records to only patients with a completed visit in 2021.

\textbf{Narrative Text Data}: We used three narrative text sources: (1) progress notes, (2) emergency department notes, and (3) narrative descriptions of social documentation from an SDOH-related module within our EHR. Progress notes typically document patient clinical status or related health events in outpatient and inpatient settings. Emergency department (ED) notes document patient care within an emergency department setting. Social history documentation is stored in our EHR as longitudinal records with the same text carried forward and edited in subsequent encounters. For simplicity of analysis and to avoid duplicate information, we only analyzed the latest social documentation records for each patient in 2021. Progress and emergency department notes were pre-processed to extract the social history sections (typically with a header of ``SOCIAL HISTORY''). Notes without this section were discarded.

\subsubsection*{Note Classification Evaluation}
Note-level classification performance was assessed by annotating 750 UW Validation Notes with five multi-class labels (one for each event type): \textit{Alcohol}, \textit{Drug}, and \textit{Tobacco} had labels $\{unknown, current, past, none\}$; \textit{Employment Status} had labels $\{unknown, employed, unemployed, retired, on\mbox{ }disability, student, homemaker\}$, and \textit{Living Status} had labels $\{unknown, alone, with\mbox{ }family, with\mbox{ }others, homeless\}$. The $unknown$ label is analogous to the $null$ label in mSpERT. Where the patient's status was described multiple times in a document, the most recent value was used. Five medical students annotated our gold standard data set. The initial annotation training round consisted of all annotators labeling the same 15 randomly selected social history sections with the extracted trigger spans from mSpERT pre-labeled. After the initial training round, 750 social history sections were single-annotated. Classification performance was evaluated using accuracy, precision (P), recall (R), and F1. The inter-rater agreement on 15 notes in the initial round of annotation was 0.95 F1. 

\subsubsection*{Structured Data and NLP Data Evaluation}

Using the extracted information and existing structured data, we assessed the proportion of patients that had a positive indication for current alcohol, tobacco, and drug use, any description of employment, and current homelessness within the one-year time period of the UW Data Set. These SDOH were selected because they provide the most direct comparison with existing structured data. In the extracted information and structured data, patients may have multiple descriptions of a given SDOH over time (e.g. alcohol use indicated as \textit{current} at one visit but \textit{past} in a subsequent visit). We counted any patient with any positive indication for listed SDOH as positive, regardless of any subsequent changes, given the short time period.

\ifsubfile
\bibliography{mybib}
\fi

\section*{Results}

\subsection*{Information Extraction}

\begin{table}[ht]
    \small
    \centering

\begin{tabular}{llrrrr}
\toprule
\textbf{Event type}           & \textbf{Argument} & \textbf{\# Gold} & \textbf{P} & \textbf{R} & \textbf{F1} \\ \midrule
\multirow{8}{*}{Substance}    & Trigger           & 1,310            & 0.94       & 0.96       & 0.95        \\
                              & Status Time       & 1,310            & 0.89       & 0.90       & 0.89        \\
                              & Amount            & 217              & 0.74       & 0.76       & 0.75        \\
                              & Duration          & 65               & 0.77       & 0.71       & 0.74        \\
                              & Frequency         & 165              & 0.73       & 0.74       & 0.73        \\
                              & History           & 103              & 0.60       & 0.69       & 0.64        \\
                              & Method            & 102              & 0.68       & 0.55       & 0.61        \\
                              & Type              & 319              & 0.76       & 0.62       & 0.68        \\ \midrule
\multirow{5}{*}{Employment}   & Trigger           & 153              & 0.94       & 0.88       & 0.91        \\
                              & Status Employ     & 153              & 0.90       & 0.84       & 0.87        \\
                              & Duration          & 5                & 0.80       & 0.80       & 0.80        \\
                              & History           & 7                & 0.80       & 0.57       & 0.67        \\
                              & Type              & 84               & 0.80       & 0.57       & 0.67        \\ \midrule
\multirow{5}{*}{Living Status} & Trigger           & 354              & 0.88       & 0.89       & 0.89        \\
                              & Status Time       & 354              & 0.87       & 0.87       & 0.87        \\
                              & Type Living       & 354              & 0.84       & 0.83       & 0.83        \\
                              & Duration          & 9                & 0.50       & 0.33       & 0.40        \\
                              & History           & 2                & 0.50       & 0.50       & 0.50        \\ \midrule
\textbf{OVERALL}              &                   & \textbf{5,066}   & \textbf{0.87}  & \textbf{0.85} & \textbf{0.86}        \\  \bottomrule
\end{tabular}
    \caption{Event extraction performance on the UW portion of the SHAC test set}
    \label{extraction_performance}
\end{table}

mSpERT was trained on the entire SHAC train set (1,316 MIMIC and 1,751 UW notes) and evaluated on the UW partition of the SHAC test set (518 notes), as the UW partition is most similar to the UW Data Set. The overall performance in Table \ref{extraction_performance} is the micro-average across all extracted phenomenon (all event types, triggers, and arguments). 
The training and test data used to develop the mSpERT SDOH extractor is identical to Subtask C of the n2c2/UW SDOH Challenge.\cite{lybarger2023n2c2} Subtask C included 10 participating teams that used a wide range of extraction approaches including pre-trained transformer-based language models (BERT\cite{devlin2019bert} and T5\cite{raffel2020exploring}). The top three teams achieved 0.89, 0.88, and 0.86 overall F1. If compared to shared task systems, the mSpERT performance of 0.86 overall F1 would be similar to the third place team.

The SHAC event structure most heavily used in the EHR case study includes the triggers and labeled arguments. The triggers resolve the event's type (\textit{Alcohol}, \textit{Drug}, etc.) and the labeled arguments capture normalized representations of important SDOH. mSpERT achieved high performance in identifying triggers for all events types (0.89-0.95 F1) and resolving the \textit{Status Time}, \textit{Status Employ}, and \textit{Type Living} multi-class labels (0.83-0.89 F1). The span-only argument (e.g. \textit{Amount}, \textit{Duration}, etc.) performance varied by argument type and event type. 

The \textit{Error Analysis} section of the Appendix includes a detailed quantitative and qualitative error analysis, focusing on triggers and labeled arguments. Substance use performance varied by event type (\textit{Alcohol}, \textit{Drug}, and \textit{Tobacco}) and \textit{Status Time} label. Across substance event types, performance was highest for the \textit{Status Time} \textit{none} label ($\ge 0.94 F1$), where descriptions tend to be relatively concise and homogeneous (e.g. ``Tobacco: denies''). Performance was lower for \textit{current} ($0.80-0.91 F1$) and \textit{past} ($0.59-0.81 F1$), which tend to be associated with more heterogeneous descriptions and have higher label confusability. Regarding \textit{Employment}, performance was relatively high for all \textit{Status Employ} labels ($\ge 0.86 F1$). The \textit{Type Living} performance was highest for \textit{with family} ($0.91 F1$) and \textit{alone} ($0.90 F1$) and lower for \textit{homeless} ($0.80 F1$) and \textit{with others }($0.69 F1$)

\subsection*{EHR Case Study}

\subsubsection*{Extractor Validation}

Table \ref{tbl_doc_level_results} presents mSpERT validation results for the note-level extraction performance on the UW Validation Notes. Precision, recall, and F1 were calculated by considering $unknown$ as the negative label, and accuracy was calculated using direct comparisons of all class labels. Comparing Table \ref{extraction_performance} to Table \ref{tbl_doc_level_results} suggests some reduction in performance associated with mapping the events extracted by mSpERT to document-level labels. However, the note-level performance is relatively high across event types (0.77-0.86 F1). 

\begin{table}[h!]
    \small
    \centering
    \def\arraystretch{1.3}
\begin{tabular}{m{1.7cm} l c c c c}
 \toprule
 
 \textbf{Category} & \textbf{Acc.} & \textbf{P} & \textbf{R} & \textbf{F1} \\
    \midrule
    Alcohol       & 0.93 & 0.88 & 0.84 & 0.86 \\
    Tobacco       & 0.92 & 0.87 & 0.85 & 0.86 \\
    Drug          & 0.94 & 0.87 & 0.87 & 0.87 \\
    Employment    & 0.86 & 0.82 & 0.72 & 0.77 \\
    Living Status & 0.87 & 0.77 & 0.80 & 0.79 \\ 
    \midrule
    OVERALL      & 0.90 & 0.84 & 0.81 & 0.83 \\
    \bottomrule
    
\end{tabular}
    \caption{Note-level performance of mSpERT on the UW Validation Notes. }
    \label{tbl_doc_level_results}
\end{table} 

\subsubsection*{Comparison of Extracted and Structured Information}

Table \ref{tbl_structured_vs_nlp} compares the extracted and structured SDOH information, including the proportions of unique patients with \textit{current} substance use (\textit{Alcohol}, \textit{Drug}, and \textit{Tobacco}), any \textit{Employment} information (\textit{employed}, \textit{unemployed}, etc.), and \textit{Living Status} of \textit{homeless}. These selections are most directly comparable between the structured data and extracted SDOH. \textit{Tobacco}, \textit{Drug}, and \textit{Living Status} showed the most significant gains in the number of patients for whom extracted SDOH revealed risk factors not captured by structured data; 32\% of homeless patients, 19\% of current tobacco users, and 10\% of current drug users only have these SDOH captured in the clinical notes without corresponding structured information. Employment showed the lowest relative gain with 11\% of patient employment found in both sources and 1\% found only by NLP. The \textit{Note Distribution} section of the Appendix presents the distribution of extracted event types by note type and provider speciality. The \textit{Alcohol and Drug usage types} section of the Appendix presents normalized past and present substance counts.

\begin{table}[ht!]
    \small
    \centering
    \def\arraystretch{1.3}
%

\newcolumntype{C}[1]{>{\centering\arraybackslash}p{#1}}

\begin{tabular}{l| C{0.82in} | C{0.22in} C{0.22in} C{0.22in}| C{0.82in} | C{0.22in} C{0.22in} C{0.22in}}
\toprule
 & \multicolumn{4}{c|}{\textbf{All Patients (N=618K)}} & \multicolumn{4}{c}{\textbf{Patients with Social History text (N=225K)}} \\
 & \multirow{2}{*}{\parbox{0.82in}{\textbf{\# Patients with SDOH info}}} & \multicolumn{3}{c|}{\textbf{SDOH Source}} & \multirow{2}{*}{\parbox{0.82in}{\textbf{\# Patients with SDOH info}}} & \multicolumn{3}{c}{\textbf{SDOH Source}} \\
 \textbf{SDOH} &   & \textbf{Struct.} & \textbf{NLP} & \textbf{Both} & & \textbf{Struct.} & \textbf{NLP} & \textbf{Both} \\
 \midrule
Alcohol (current)        & 148,221 & 87\% & 5\%  & 8\%  & 106,658 & 82\% & 7\%  & 11\% \\
Tobacco  (current)       & 34,871  & 69\% & 19\% & 12\% & 25,675  & 56\% & 26\% & 17\% \\
Drug  (current)          & 42,309  & 86\% & 10\% & 5\%  & 31,222  & 81\% & 13\% & 6\% \\
Employment (any status)  & 575,278 & 88\% & 1\%  & 11\% & 200,926 & 66\% & 4\%  & 30\% \\
Living Status (homeless) & 11,567  & 64\% & 32\% & 4\%  & 9,390   & 55\% & 39\% & 5\% \\

\bottomrule    
\end{tabular}
    \caption{Proportions of unique patients with current SDOH found only in structured data, only by NLP extraction, or in both. For purposes of comparison, in the right-most three columns we further limit structured data to only patients who had social history narrative text as well.}
    \label{tbl_structured_vs_nlp}
\end{table}

\section*{Discussion}

Our SDOH extraction approach provides a promising way to identify patients' SDOH and social needs. We demonstrate high performance, especially in identifying SDOH events (triggers) and determining status and type labels. The SDOH from prior work that is most comparable to our EHR case study is tobacco use. 19\% of potential smokers identified were found only by NLP, and Navarthe et al. similarly identified 15\% of patients, though only among cardiovascular disease patients and as compared to ICD-9 codes.\cite{navathe2018hospital} Yu et al.'s study of cancer cohorts similarly found 18\% of smoking information using only NLP.\cite{yu2021study} Our findings differed from Wang et al.'s, who found 52\% of a small cohort with smoking habits using NLP, but all of whom also had corresponding structured data.\cite{wang2016investigating} This may be due to differing institutional practices or other confounding factors.  

Unlike previous studies which extracted SDOH using text classification or NER, our detailed SDOH representation may better aid clinicians in identifying SDOH documented in notes by determining chronicity, duration, frequency, and type. This event-based approach can  automatically generate detailed summaries of patient SDOH risk factors, reducing clinician chart review time. 
While EHRs offer structured fields to document social needs, the consistency of this information collection depends on competing priorities.\cite{berg2022automating} Given the expanding quantity of EHR data, it is increasingly important for clinicians to efficiently identify key information that informs patient care, including SDOH and social needs. NLP serves to bridge the gap between unstructured clinical narratives and structured data by augmenting existing structured data and identifying otherwise unknown social needs. Our study explores the entire patient population at a health system in an urban setting. Our findings may be most generalizable to other urban hospital systems; however, we leave this examination to future work.

Developing SDOH extraction capabilities is timely, given new guidelines released by the Center for Medicare \& Medicaid Services that will request screening rates for SDOH and social needs in 2023 and require reporting in 2024.\cite{IPPS} Our investigation indicates important SDOH information can be extracted from the clinical narrative with high performance to augment structured data.

\subsection*{Limitations and Future Work}

While we extracted SDOH from a large clinical data set spanning the UW medical system, our investigation only used progress notes, emergency notes, and social history text, which are a subset of documentation and likely do not represent all documented SDOH. Our EHR case study was limited to data from one year, and the performance of mSpERT for other time periods or institutions is not well understood. The prevalence and patterns of SDOH descriptions in narrative text may vary over time and by institution. 

This study is limited by extractor performance and target SDOH. Although performance was relatively high for most SDOH information, extraction errors negatively impact the case study, and certain SDOH will be disproportionately affected. For example, substance abstinence was extracted with higher performance than current or past substance use. The SHAC annotations and this study capture substance use, employment, and living status information; however, there are many important SDOH that are not addressed, such as living environment, access to care, and food security.

Future studies are needed to understand how extracted SDOH should be incorporated into social needs screening. Topics of interest could include methods for integration into the medical record and reducing the need for manual data entry,\cite{bakken2019can} impact of false positives on stigmatization,\cite{Hartzler2023integrating} and influence on patient access to healthcare or social services.

\ifsubfile
\bibliography{mybib}
\fi

\section*{Conclusions}

SDOH are increasingly recognized for their impact on patient well-being and public health. The clinical narrative contains rich descriptions of SDOH, and the automatic extraction of SDOH from these narratives can enable large-scale use of the information they contain. We introduce a multi-label version of the entity and relation extraction SpERT architecture, mSpERT, which can extract overlapping spans (entities) and assign multiple labels to spans. mSpERT achieves high performance on the UW partition of the SHAC test set at 0.86 F1 overall. mSpERT achieves especially high performance for event (trigger) identification (0.89-0.95 F1) and the status and type arguments (0.83-0.89 F1) that characterize the most salient aspects of the SDOH.

In an EHR case study, we processed 430,406 free-text descriptions of SDOH using mSpERT and automatically compared the extracted structured semantic representations of SDOH to existing structured EHR data. Based on our analysis, combining the narrative SDOH information with the existing structured data yields a more comprehensive patient representation that can be used to guide patient care, assess health risks, and identify social needs.

\ifsubfile
\bibliography{mybib}
\fi

\section*{Acknowledgements}
This work was done in collaboration with the UW Medicine Analytics Department.

\section*{Funding Statement}
This work was supported in part by the National Institutes of Health (NIH) - National Cancer Institute (Grant Nr. R21CA258242-01S1),  NIH - National Library of Medicine (NLM) Biomedical and Health Informatics Training Program at the University of Washington (Grant Nr. T15LM007442), NIH - National Center for Advancing Translational Sciences (NCATS) (Institute of Translational Health Sciences, Grant Nr. UL1 TR002319). The content is solely the responsibility of the authors and does not necessarily represent the official views of the National Institutes of Health. 

\section*{Competing Interests Statement}
The authors have no competing interests to declare.

\section*{Contributorship Statement}
KL and NJD are co-lead authors for the manuscript and contributed equally. All authors contributed to the study design. KL, NJD, and RL developed and implemented algorithms and analyzed the data. KL and NJD drafted the initial manuscript. All authors contributed to the interpretation of the data, manuscript revisions, and intellectual value to the manuscript.

\section*{Data Availability Statement}
The SHAC data set used in this work will be made available through the University of Washington.

\bibliography{mybib}

\pagebreak

\section*{Figure Legends}
\subsection*{Manuscript body}
\noindent
Figure 1: SHAC annotation example (left side) with slot filling interpretation of the annotated events (right side). \\

\noindent
Figure 2: Multi-label Span-based Entity and Relation Transformer (mSpERT) model, which builds on the original SpERT framework.\cite{eberts2020span} \\


\subsection*{Manuscript appendix}



\noindent
Figure 3: Performance breakdown by argument subtype label. The left-hand y-axis is the micro-averaged F1 for the argument subtype labels (vertical bars). The right-hand y-axis is the number of gold events (\ding{58}).

Figure 4. Counts of unique patients found from normalized results of the most common drug and alcohol types extracted from free-text. Normalized values include both non-specific categories, such as ``illicit'' and ``recreational'', as well as specific drug and alcohol types. We found a total of 5,807 patients with normalized drug use values, of whom 1,503 (25\%) were found to use more than one drug. A total of 1,429 patients had normalized alcohol use values, of whom 119 (8\%) were found to use more than one kind of alcohol. 225 patients had normalized values for both drugs and alcohol.

\pagebreak

\section*{Appendix}

\subsection*{Error Analysis}

 Figure 3
 \begin{figure}[!ht]
     \centering
     \frame{\includegraphics[width=5.0in]{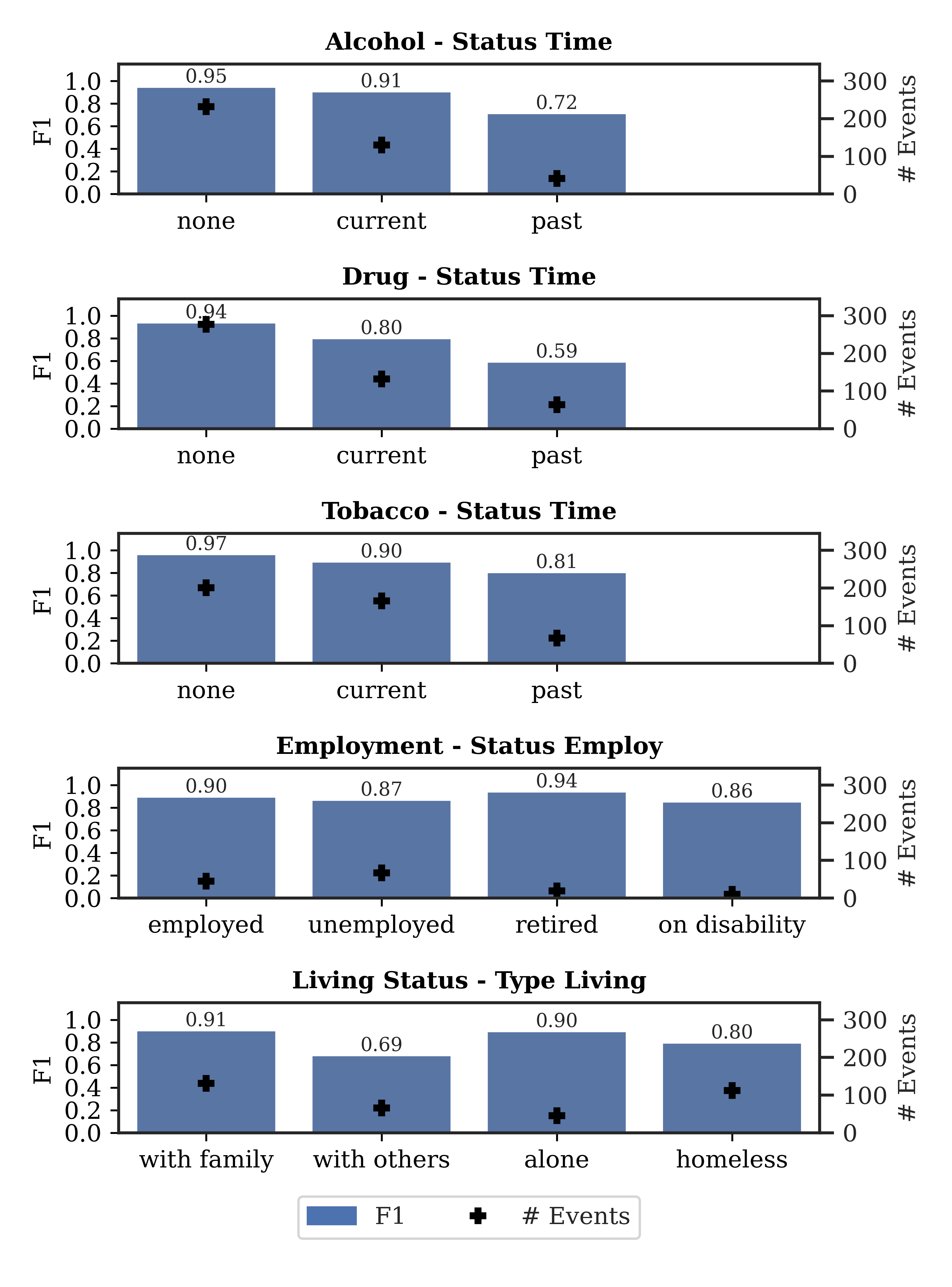}}
     \caption{Performance breakdown by argument subtype label. The left-hand y-axis is the micro-averaged F1 for the argument subtype labels (vertical bars). The right-hand y-axis is the number of gold events (\ding{58}).}
     \label{error_analysis}
 \end{figure}

To better understand the performance of the trained mSpERT SDOH extractor, we performed an error analysis, including a quantitative analysis of the labeled arguments and a qualitative analysis of the social history sections with lower performance. The EHR case study relies most heavily on the trigger and labeled argument predictions, so the error analysis focuses on these phenomena. This error analysis is based on the predictions of the extractor on the SHAC UW test set (same evaluation data as Table \ref{extraction_performance}). 

Figure 3 presents the performance for the labeled arguments, which capture normalized representations of key SDOH. The performance is presented for: i) \textit{Status Time} labels for \textit{Alcohol}, \textit{Drug}, and \textit{Tobacco}), ii) \textit{Status Employ} labels for \textit{Employment}, and iii) \textit{Type Living} labels for \textit{Living Situation}. The \textit{student} and \textit{homemaker} labels for \textit{Status Employ} are omitted, due to low frequency. For substance use, the \textit{none} label has the highest performance and is associated with relatively concise and heterogeneous descriptions (e.g. ``Tobacco use: none''). Performance is lower for the \textit{current} and \textit{past} labels, which tend to have more linguistically diverse descriptions and higher label confusability. Performance is lower for the \textit{Drug} \textit{current} and \textit{past} labels than for \textit{Alcohol} and \textit{Tobacco} because of the more heterogeneous descriptions of drug use (e.g. ``cocaine remote,'' ``smokes MJ,'' and ``injects heroin''). For \textit{Employment}, performance is relatively high for all subtype labels. The \textit{retired} label has the highest performance, due to the frequent use of the phrase ``retired.'' The \textit{on disability} label has the lowest performance, which is attributable to annotation and classification ambiguity associated with the presence of a disability versus receiving disability benefits. For \textit{Living Status}, higher performance is achieved for \textit{with family} and \textit{alone}, and lower performance is achieved for \textit{homeless} and \textit{with others}. The lower performance for the \textit{homeless} label is attributable to the linguistic diversity of the housing instability descriptions (see qualitative error analysis below for examples), including references to named homeless shelters. The \textit{with others} label is applicable to a wide range of living situations, including living with friends, non-married partners, skilled nursing facilities, adult family homes, hospitals, etc., which contributes to the lower extraction performance.

Contributing factors to false negatives include misspellings (e.g. ``not currently workign''), uncommon abbreviations (e.g. ``tobb'' for tobacco), and linguistic diversity. False negatives related to linguistic diversity impact all event types but are particularly prominent for \textit{Employment} and \textit{Living Status}. For \textit{Employment} events, there are often employment related keywords, like ``works'' or ``unemployed''; however, there are some descriptions of employment, where employment must be inferred from specific jobs, for example ``She is an artist and a former RN.'' For \textit{Living Status}, there are diverse descriptions of housing instability (\textit{homeless} label), for example ``Homeless,'' ``Currently staying in an RV,'' or ``Undomicilled, bouncing from house to house.'' The predictions also include false-positives associated with note structure and formatting. Examples include blank substance use templates, like ``Alcohol: \_,'' and atypical subheadings, like the use of ``Recreational: marijuana'' instead of ``Drug Use: Occasional marijuana.'' The predictions include false positives associated with living situation descriptions, like ``Lives in his own home in [LOCATION].,'' that lack sufficient information to resolve the \textit{Type Living} label, a necessary condition for a \textit{Living Status} event.

In the UW test set, most SDOH descriptions in a note can be represented through a single event; however, more nuanced descriptions of SDOH require multiple events to capture the available information and tend to be more challenging annotation and classification targets. These more nuanced SDOH include descriptions of: i) prior and current substance use habits or polysubstance use, ii) conflicting patient-reported and EHR-documented substance use information, iii) multiple job transitions, including employment instability, and iv) multiple recent housing transitions, including cycling in and out of homelessness. As an example, the extractor achieved low performance for the sentence, ``Drug Use: Does not endorse drug use, polysubstance use listed in chart, recent U tox showed meth, cannabinoids, opiates.'' This sentence contains multiple frequently annotated Drug triggers (``Drug Use,'' ``polysubstance use'', and the specific drug types). It also contains conflicting drug use descriptions with the patient denying drug use and a recent urine toxicology report supporting polysubstance use. The extraction performance tends to decrease, as the number events needed to represent an SDOH event type (\textit{Alcohol}, \textit{Drug}, etc.) increases.

\subsection*{Note Distribution}

Additional detailed results of our EHR case study are shown in Tables \ref{tbl_triggers_by_note_type} and \ref{tbl_triggers_by_specialty_type}. Table \ref{tbl_triggers_by_note_type} shows the proportions of note types for each SDOH, with progress notes being the majority for each SDOH, but a significant proportion of drug extractions from ED notes (32\%) and employment and living status extractions from social history documentation (47\% and 38\%). Table \ref{tbl_triggers_by_specialty_type} shows the proportions of progress notes with social history sections and proportion of each SDOH stratified by provider specialty.

\begin{table}[ht]
    \small
    \centering
    \def\arraystretch{1.3}
\begin{tabular}{m{1.7cm} c| c c c c}
 \toprule
 
 \textbf{SDOH} & \textbf{Total Notes} & \multicolumn{3}{c}{\textbf{SDOH extractions by note type}} \\
  & & \textbf{Progress Notes} & \textbf{ED Notes} & \textbf{Social History Doc.} \\
    \midrule
    Alcohol       & 41,623  & 32,559 (78\%) & 7,544 (18\%) & 1,520 (4\%) \\
    Tobacco       & 68,573  & 60,413 (88\%) & 6,402 (9\%)  & 1,758 (3\%) \\
    Drug          & 28,715  & 17,872 (62\%) & 9,321 (32\%) & 1,522 (5\%) \\
    Employment    & 129,952 & 64,824 (50\%) & 3,531 (3\%) & 61,597 (47\%) \\
    Living Status & 66,992  & 36,904 (55\%) & 4,410 (7\%) & 25,678 (38\%) \\ 
    \bottomrule
    
\end{tabular}

    \caption{Counts of unique free-text documents with SDOH extracted stratified by SDOH and note type.}
    \label{tbl_triggers_by_note_type}
\end{table}

\begin{table}[!h]
    \small
    \centering
    \def\arraystretch{1.3}
\begin{tabular}{l c c| c c c c c}
 \toprule
 
 \textbf{Specialty} & \textbf{\# Notes} & \textbf{\# Notes with} & \multicolumn{5}{c}{\textbf{\% Notes with SDOH}} \\
  & & \textbf{Social History} & \textbf{Alcohol} & \textbf{Tobacco} & \textbf{Drug} & \textbf{Employment} & \textbf{Living Status} \\
    \midrule

    Family Practice & 549,949 & 25,628 (5\%) & 0.8\% & 0.6\% &0.1\% & 0.8\% &0.2\%\\
    Dermatology & 83,081 & 20,121 (24\%) & 0.2\% & 8.3\% &0.2\% & 9.6\% &1.8\%\\
    Internal Medicine & 243,840 & 18,733 (8\%) & 0.7\% & 1.0\% &0.5\% & 1.8\% &1.6\%\\
    Ophthalmology & 82,007 & 17,350 (21\%) & 1.0\% & 8.4\% &0.9\% & 3.8\% &0.6\%\\
    Neurology & 62,852 & 10,848 (17\%) & 0.9\% & 1.6\% &0.6\% & 7.0\% &2.9\%\\
    Rheumatology & 35,568 & 10,776 (30\%) & 1.8\% & 10.0\% &0.6\% & 5.0\% &1.2\%\\
    Cardiology & 50,960 & 9,771 (19\%) & 3.6\% & 5.4\% &3.7\% & 3.8\% &1.0\%\\
    Hematology/Oncology & 51,912 & 9,416 (18\%) & 2.3\% & 2.4\% &0.5\% & 6.1\% &4.1\%\\
    Obstetrics/Gynecology & 118,988 & 8,571 (7\%) & 1.2\% & 1.8\% &0.4\% & 1.2\% &0.4\%\\
    Otolaryngology & 46,855 & 8,005 (17\%) & 3.2\% & 5.1\% &0.7\% & 2.2\% &0.2\%\\
    Physical Med \& Rehab & 38,855 & 7,820 (20\%) & 1.0\% & 5.7\% &0.5\% & 4.3\% &3.5\%\\
    Gastroenterology & 36,193 & 7,470 (21\%) & 1.9\% & 3.1\% &4.3\% & 6.2\% &1.8\%\\
    Psychiatry & 41,160 & 7,460 (18\%) & 0.4\% & 0.3\% &0.3\% & 7.1\% &5.4\%\\
    Orthopedic Surgery & 70,216 & 6,768 (10\%) & 3.7\% & 2.1\% &0.3\% & 1.6\% &0.3\%\\
    Sports Medicine & 51,382 & 6,608 (13\%) & 1.8\% & 5.1\% &0.3\% & 4.0\% &0.4\%\\
    Endocrinology & 30,205 & 6,404 (21\%) & 6.9\% & 4.3\% &1.0\% & 4.2\% &2.2\%\\
    Urology & 66,083 & 3,987 (6\%) & 0.7\% & 3.8\% &0.2\% & 0.8\% &0.2\%\\
    Nephrology & 25,602 & 3,472 (14\%) & 0.7\% & 2.2\% &3.1\% & 3.7\% &2.8\%\\
    Cardiac Electrophysiology & 8,540 & 2,628 (31\%) & 2.2\% & 8.9\% &6.1\% & 6.9\% &2.0\%\\
    Geriatric Medicine & 8,782 & 1,867 (21\%) & 3.0\% & 1.0\% &5.8\% & 2.6\% &6.3\%\\
    \bottomrule
    
\end{tabular}

    \caption{Counts of unique progress notes with SDOH extracted stratified by SDOH and provider specialty, limited to the top 20 specialities in terms of social history section count.}
    \label{tbl_triggers_by_specialty_type}
\end{table}

\subsection*{Alcohol and Drug usage types}

We used the extracted SDOH information to better understand the prevalence of specific substances by  use status. Structured records related to drug and alcohol use do not specify the substance types used, while the clinical narratives often include this information. To categorize the extracted drug types, we used regular expressions to normalize extracted spans. Results are shown in Figure 4. Marijuana and methamphetamines were the most frequently mentioned drugs, while wine and beer were frequently mentioned as types of alcohol used but notably less often than drug types.

 Figure 4
 \begin{figure*}[h!]
   \begin{center}
     \includegraphics[scale=1]{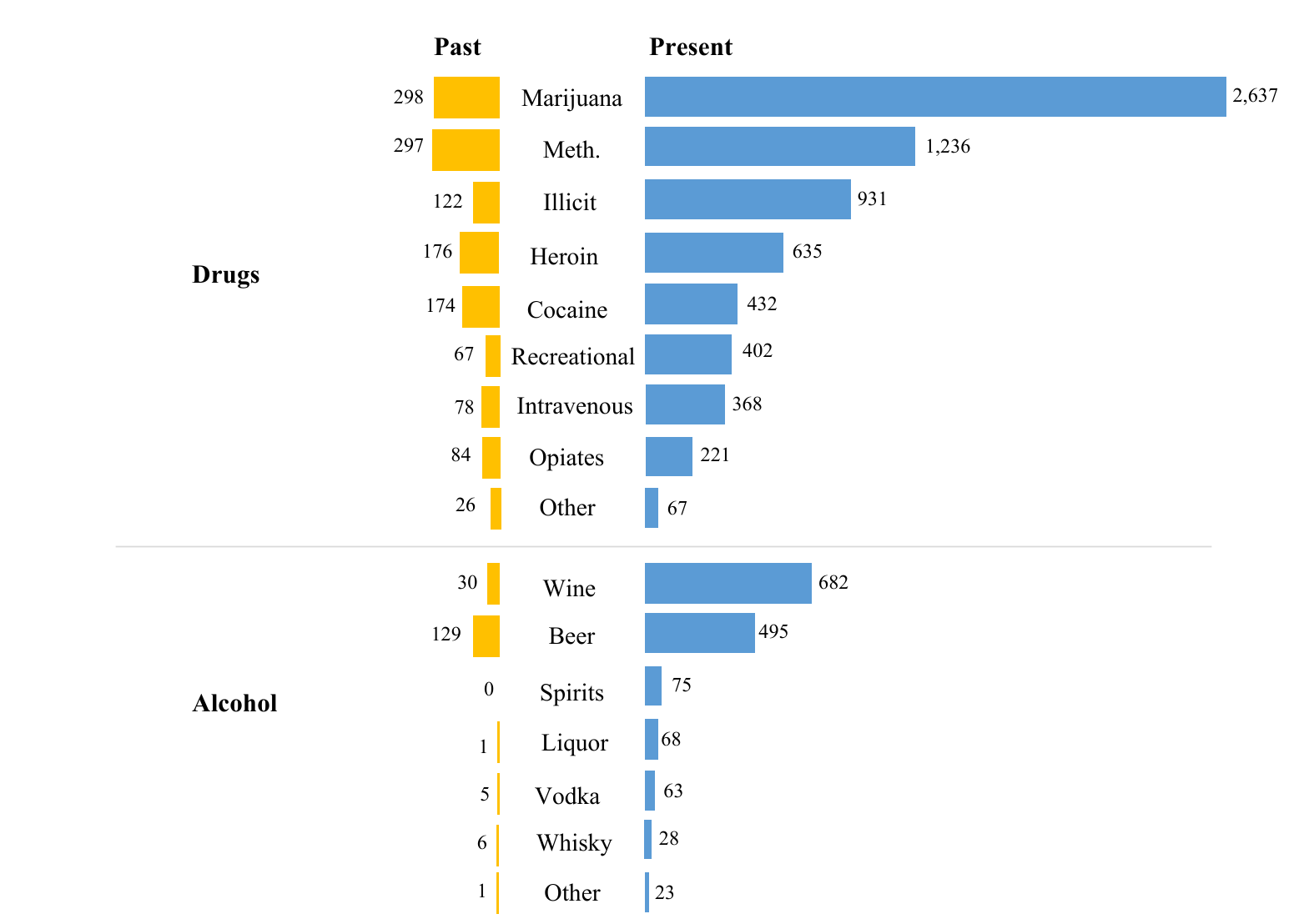}
   \end{center}
   \caption{Counts of unique patients found from normalized results of the most common drug and alcohol types extracted from free-text. Normalized values include both non-specific categories, such as ``illicit'' and ``recreational'', as well as specific drug and alcohol types. We found a total of 5,807 patients with normalized drug use values, of whom 1,503 (25\%) were found to use more than one drug. A total of 1,429 patients had normalized alcohol use values, of whom 119 (8\%) were found to use more than one kind of alcohol. 225 patients had normalized values for both drugs and alcohol.}
   \label{fig_drug_alcohol}
 \end{figure*}

\ifsubfile
\bibliography{mybib}
\fi



\end{document}